\documentclass[journal,comsoc]{IEEEtran}
\usepackage{amsmath,amsfonts}
\usepackage{algorithmic}
\usepackage{algorithm}
\usepackage{array}
\usepackage[caption=false,font=normalsize,labelfont=sf,textfont=sf]{subfig}
\usepackage{textcomp}
\usepackage{stfloats}
\usepackage{url}
\usepackage{verbatim}
\usepackage{graphicx}
\usepackage{cite}
\usepackage[utf8]{inputenc}
\usepackage[english]{babel}
\usepackage{multirow}
\usepackage[backref]{hyperref}

\begin{document}

\title{Semantic Similarity Score for Measuring Visual Similarity at Semantic Level}

\author{Senran Fan, Zhicheng Bao, Chen Dong*, Haotai Liang, Xiaodong Xu, \textit{Senior Member}, \textit{IEEE},  Ping Zhang, \textit{Fellow}, \textit{IEEE}
\thanks{Senran Fan, Zhicheng Bao, Haotai Liang, Xiaodong Xu and Ping Zhang are with the State Key Laboratory of Networking and Switching Technology, Beijing University of Posts and Telecommunications, Beijing, 100876, China. (E-mail: FSR@bupt.edu.cn; zhicheng$\_$bao@bupt.edu.cn; lianghaotai@bupt.edu.cn; xuxiaodong@bupt.edu.cn; pzhang@bupt.edu.cn)}
\thanks{*Chen Dong is the corresponding author and with the State Key Laboratory of Networking and Switching Technology, Beijing University of Posts and Telecommunications, Beijing, 100876, China. (E-mail: dongchen@bupt.edu.cn)}
\thanks{X. Xu and P. Zhang are also with the Department of Broadband Communication, Peng Cheng Laboratory, Shenzhen, China.}}

\markboth{}
{}

\maketitle

\begin{abstract}
    Semantic communication is considered a promising novel communication paradigm. Unlike traditional symbol-based error-free communication systems, semantic-based visual communication systems extract, transmit, and reconstruct images at the semantic level. However, widely used image similarity evaluation metrics, PSNR or MS-SSIM, struggle to accurately measure the loss of semantic-level information during transmission. This presents challenges in evaluating the performance of visual semantic communication systems, especially when comparing them with traditional systems. To address this, we propose a semantic evaluation metric---SeSS (Semantic Similarity Score), based on Scene Graph Generation and graph matching, which shifts the similarity scores between images into graph matching scores. Meanwhile, similarity scores for tens of thousands of image pairs are manually annotated to fine-tune the hyperparameters in the SeSS, aligning SeSS more closely with human semantic perception. The performance of the SeSS is tested on different datasets, including (1)traditional and semantic communication systems at different compression rates, (2)communication systems at different signal-to-noise ratios, (3)large-scale model with different noise levels introduced, and (4)cases of images subjected to certain special transformations. The experiments demonstrate the effectiveness of SeSS, indicating that SeSS can measure the semantic-level differences in semantic-level information of images and can be used for evaluation in visual semantic communication systems.
\end{abstract}


\section{Introduction}
\IEEEPARstart{W}ith the increasing demand for intelligent services, 6G wireless networks will evolve from the traditional architecture to a new architecture based on widespread smart connectivity. As mentioned in \cite{semantic}, as a novel architecture, semantic communication comprehensively considers the semantic information involved in user requirements, application needs, and data transmission processing, and is expected to become a bright paradigm of communication systems.\par
Semantic communication was first proposed in Shannon and Weaver's paper\cite{shannon}, where they divided communication into three levels: technical-level, semantic-level, and effectiveness-level. According to \cite{6G}, semantic communication systems no longer require error-free transmission at the symbolic level. Instead, they improve system efficiency by extracting, transmitting, and reconstructing the semantic information in the source. The operation of semantic communication systems focuses on the semantic level, with the communication process aiming to preserve the semantic information in the source, while ensuring similarity or consistency at the semantic level between the source and the destination. Therefore, when evaluating semantic communication systems, it is reasonable to measure the loss of semantic information in the source and destination. However, as semantic communication technology is still in the development stage, research on evaluation metrics for semantic communication systems is relatively limited, particularly the lack of structured, interpretable image similarity evaluation metrics at the semantic level.\par
Existing image similarity metrics can be categorized into pixel-level (e.g. PSNR/MSE), structure-level (e.g. SSIM/MS--SSIM\cite{ssim}), and semantic-level (e.g. LPIPS\cite{lpips}/ViTScore\cite{vitscore}/ClipScore\cite{clipscore}). As shown in Fig. 1, these metrics progress from focusing on low-level pixel differences to high-level semantic information. Metrics on the left side, while sensitive to low-level pixel/structure differences, have limited "understanding" of image semantics and cannot accurately measure semantic-level information loss. In contrast, the metrics on the right focus more on high-level semantic differences, with higher tolerance for low-level variations. However, current semantic-level metrics also have limitations. LPIPS and ViTScore relies on convolutional features and have difficulties in capturing holistic semantic information. ClipScore, while incorporating global context, suffer from issues like lack of structured design and interpretability, as well as difficulties in accurately understanding object relationships\cite{VAQscore}. Therefore, our goal is to develop a structured, interpretable image similarity metric that can effectively quantify semantic-level differences between images.\par
As mentioned in \cite{knowledge}, information is the carrier of knowledge, and the core idea of semantic communication is that humans can infer and extract information from signals using their prior knowledge. Entering the visual domain, human cognition of complex images often involves the recognition of objects and the inference of relationships between them. It’s believed that by referencing this process, image similarity metrics can be made to better "understand" the semantic-level information in images, aligning with human perception, and also endowing the metrics with structured features and interpretability. Therefore, Scene Graph Generation (SGG) technology\cite{sgg} is considered in this work. Scene graphs provide a structured representation of a scene, clearly expressing the objects, attributes, and relationships between objects. Scene Graph Generation techniques based on deep learning, utilizing object detection, cross-modal analysis and other technologies, can achieve higher-level understanding and reasoning of visual scenes, and convert images into structured object-relationship graphs. \par
Based on the above research background, the image similarity metric SeSS is proposed. In this metric, the image is first generated into an object-relation graph through Scene Graph Generation technology, and then these graphs are combined with the CLIP model and graph matching algorithm to finally calculate the similarity score between the images. With SGG technology as the bridge, the calculation of semantic similarity between images is transformed into the graph matching score calculation between the corresponding graphs of the images. This makes the metric constrained to "understand" and compare the differences between images in a way that aligns with the semantic level and human perception.\par
The main contributions of this paper are summarized as follows.

\begin{figure*}[ht]
    \centering
    \includegraphics[scale=0.55]{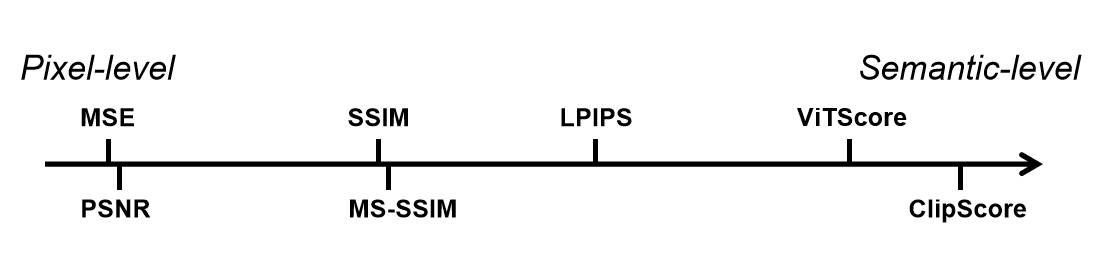}
    \caption{The existing image similarity metrics can be arranged according to the information level they focus on, with the metrics on the left focusing more on structure and pixel-level differences between images, and the metrics on the right focusing more on semantic-level differences between images. Among these, MSE and PSNR are typical pixel-level metrics, SSIM and MS-SSIM are structure-level metrics, while LPIPS, ViTScore and ClipScore can be considered semantic-level metrics.}
\end{figure*}  

\begin{itemize}
    \item[(1)]
A novel image similarity metric called SeSS is proposed. SeSS is based on SAM segmentation model, Scene Graph Generation (SGG) technology, and graph matching. The computation process of SeSS is both structured and interpretable, and it can measure the semantic-level differences between images.
\end{itemize}

\begin{itemize}
    \item[(2)]
Semantic similarity scores between 100,000 image pairs are manually annotated, and used to tune the hyperparameters in the graph matching algorithm of the proposed metric SeSS, making it better align with human semantic perception.
\end{itemize}

\begin{itemize}
    \item[(3)]
Experiments demonstrate the performance of the proposed SeSS. SeSS is compared with 6 typical metrics (PSNR, SSIM, MS-SSIM, LPIPS, ViTScore, ClipScore) in 4 types of experiments: (1) Images transmitted by traditional and semantic communication image systems under different compression ratios. (2) Images transmitted by traditional and semantic communication image systems under different SNRs. (3) Synthetic similar images generated by large models with different noises introduced. (4) Special transformed image cases. The experiments show that SeSS has good performance and robustness, and can effectively evaluate the performance of visual semantic communication systems.
\end{itemize}

This paper is arranged as follows. In section II, the existing image similarity metrics are introduced. In section III, the complete framework and the specific algorithm of the proposed image semantic similarity metric SS are introduced. In section IV, the experimental data and visual examples from various experiments are provided to demonstrate the effectiveness of the SeSS. Finally, conclusions of this paper are drawn in section V.
\section{Related Work}

Image similarity metrics are used to evaluate the differences between images, providing important performance evaluation benchmarks for visual communication systems\cite{sc1, sc2, lsci, sc3, mdvsc}. The widely used image similarity metrics like PSNR or SSIM are still based on pixel or structure, struggling to measure the loss at the semantic level. The lack of semantic-level image similarity metrics forces the design of visual semantic communication systems to retain a certain amount of non-semantic information in order to achieve better evaluation metrics.\par
For semantic communication systems that leverage semantic-level information to achieve extreme image compression, although the transmitted images may have higher visual or semantic similarity to the original, the loss of non-semantic information still hinders them from outperforming traditional communication systems in terms of PSNR or SSIM metrics. This situation hinders the development of semantic communication.\par 
As shown in Table I, the existing metrics are categorized into three types, with a detailed introduction to their working principles, advantages, and limitations in evaluating visual semantic communication systems.\par
~\\

\begin{table*}[ht]
    \renewcommand\arraystretch{1.7}
    \begin{flushleft}   
    \caption{Comparison between Image Similarity Metrics} 
	\centering
    \label{table:1} 
	\begin{tabular}{|m{1.2cm}|m{1.2cm}|m{5.6cm}|m{5.6cm}|m{2.0cm}|} 
        \hline Metric & Level & Advantages & Key Limitations & Interpretability \\ 
		\hline MSE & \multirow{2}*{Pixel} & 1.Physically Meaningful &  1.All Bits are Considered Equally Important & High\\ 
        \cline{1-1} \cline{5-5} PSNR &   & 2.Easy to Compute and Widely Used & 2.Exhibits Large Discrepancy with Human Perception of Image Differences & High\\
        \hline SSIM & \multirow{2}*{Structure} & 1.Sensitive to Changes in Luminance, Contrast, and Structure &  1.Struggles to Capture Semantic Information in Images & High\\ 
        \cline{1-1} \cline{5-5} MS-SSIM &   & 2.Easy to Compute and Widely Used & 2.Not Robust to Changes Weakly Correlated with Image Semantics & High\\
        \hline LPIPS & \multirow{4}*{Semantic} & 1.DL-based Metrics give more attention to Semantic-leval information &  Struggles to Capture semantic information of the entire image & Weak\\ 
        \cline{1-1} \cline{4-5} ViTScore &   & 2.More Consistent with Human Perception about Image Differences  & Not Robust to Changes Weakly Correlated with Image Semantics & High\\
        \cline{1-1} \cline{4-5} ClipScore &   & & Prone to Confusing the Subject and Object in Relationships & Weak\\
        \cline{1-1} \cline{3-5} SeSS & & 1.Structured and Interpretable  \par  2.Robust to Changes Weakly Correlated with Image Semantics & Imprecise in Processing Images with Weak Semantic Content, e.g. noise patterns or color blocks & High\\
        \hline
	\end{tabular}
\end{flushleft} 
\end{table*}

\noindent \emph{A. Pixel-level Metrics}\par
MSE is the simplest and most commonly used image similarity metric. Given two images of the same size, \mbox{$X$} and \mbox{$Y$}, where the pixel values are denoted as \mbox{$x_i$} and \mbox{$y_i$} respectively, and the total number of pixels is \mbox{$N$}, the MSE can be expressed as:\par
\begin{equation}\label{eqn-1} 
    \text{MSE}(X,Y) = \frac{1}{N}\sum_{i=1}^{N}{(x_i - y_i)}^2.
\end{equation}
PSNR is to convert MSE into the commonly used decibel form in signal processing. From a digital perspective, it has more distinguishing power compared to MSE. Its calculation is based on MSE:\par
\begin{equation}\label{eqn-1} 
    \text{PSNR}(X,Y) = 10lg{\frac{L^2}{\text{MSE}(X,Y)}}.
\end{equation}

Both of these metrics are obtained by performing per-pixel calculations between images. Their advantage is that they have a clear physical definition and are easy to calculate, which is why they are widely used. However, they treat all pixels in the image as equally important, and therefore completely fail to "understand" the differences in semantic-level information.\par
~\\
\emph{B. Structure-level Metrics}\par
The structure-level metrics SSIM and MS-SSIM have proposed improvements by involving The local structural information of the image. They no longer focus on isolated pixels, but instead calculate by combining the pixels in a region. The specific algorithm is as follows,\par
\begin{equation}\label{eqn-1} 
    l(x,y) = \frac{2\mu_x\mu_y+C_1}{\mu_x^2+\mu_y^2+C_1}.
\end{equation}
\begin{equation}\label{eqn-1} 
    c(x,y) = \frac{2\sigma_x\sigma_y+C_2}{\sigma_x^2+\sigma_y^2+C_2}.
\end{equation}
\begin{equation}\label{eqn-1} 
    s(x,y) = \frac{\sigma_{xy}+C_3}{\sigma_x\sigma_y+C_3}.
\end{equation}
\begin{equation}\label{eqn-1} 
    \text{SSIM}(x,y) = l(x,y)^{\alpha}c(x,y)^{\beta}s(x,y)^{\gamma}.
\end{equation}
Where \mbox{$l(x,y)$} is the luminance similarity, \mbox{$c(x,y)$} is the contrast similarity, and \mbox{$s(x,y)$} is the structure similarity. Therefore, the SSIM metric is sensitive to the changes in these three aspects. The SSIM metric calculates the similarity of each window block, and the final metric is obtained by averaging the metrics of all blocks. MS-SSIM, on the other hand, calculates the average similarity between images by using calculation windows of different scales.\par
\begin{equation}\label{eqn-1} 
    \text{MS-SSIM}(x,y) = l_{M}(x,y)^{\alpha_M} \prod_{j=1}^{M}c_j(x,y)^{\beta_j}s_j(x,y)^{\gamma_j}.
\end{equation}
Where \mbox{$M$} is the number of different window blocks used. However, regardless of how many window blocks are used, due to the limited calculation view based on the windows, the structure-level metrics find it difficult to "understand" the global semantic information. \par
~\\
\emph{C. Semantic-level Metrics}\par
It is only by understanding the semantic information of regions and the entire image at a higher level, and comprehensively considering it, that the difference in information can be evaluated at the semantic level. This is beyond the scope of traditional algorithms. With the development of deep learning technology, it has become possible to extract the semantic information of images through neural networks and perform difference calculations.\par
LPIPS is the most commonly used, deep learning-based image similarity metric, but it has not yet fully reached the semantic level. Rather, it makes judgments more in line with human perception through data-driven convolutional neural networks.\par
\begin{equation}\label{eqn-1} 
    d(x,x_0) = \sum_l \frac{1}{H_lW_l} \sum_{h,w}||\omega_l \odot (\hat{y}_{hw}^l-\hat{y}_{0hw}^l)||_2^2.
\end{equation}
\begin{equation}\label{eqn-1} 
    \text{LpipsScore}(x,x_0) = 1 - d(x,x_0).
\end{equation}
LPIPS evaluates the similarity between two images by taking the average pooling of the feature distance at corresponding positions, which is no longer heavily influenced by low-level information.\par
ViTScore  uses the Vision Transformer (ViT)\cite{vit} architecture to introduce the contextual relationships and global information of the image for the calculation. ViTScore extracts the feature vectors of each patch obtained by ViT before the classification layer and uses these feature vectors to calculate the similarity between images,\par
\begin{equation}\label{eqn-1} 
    ViT(X) = (x_0, x_1, ..., x_{n-1}).
\end{equation}
\begin{equation}\label{eqn-1} 
    ViT(Y) = (y_0, y_1, ..., y_{m-1}).
\end{equation}
\begin{equation}\label{eqn-1} 
    R_{ViTScore}(X,Y) = \frac{1}{n} \sum_{i=0}^{n-1} \max_{0 \leq j < m} x_i^T y_j.
\end{equation}
\begin{equation}\label{eqn-1} 
    P_{ViTScore}(X,Y) = \frac{1}{m} \sum_{j=0}^{m-1} \max_{0 \leq i < n} x_i^T y_j.
\end{equation}
\begin{equation}\label{eqn-1} 
    \text{ViTScore}(X,Y) = 2\frac{R_{ViTScore}(X,Y) \cdot P_{ViTScore}(X,Y)}{R_{ViTScore}(X,Y) + P_{ViTScore}(X,Y)}. 
\end{equation}
Its calculation process is to match the most semantically similar patches between the input images \mbox{$X$} and \mbox{$Y$}, and then calculate the average matching score. \par
ClipScore is based on the multimodal model CLIP\cite{clip}, which aligns images and natural language descriptions, enabling it to "understand" images at a higher level. As a result, it is the most robust to changes weakly correlated with image semantics. Therefore, it has been widely used in research fields related to visual semantic communication.\par
\begin{equation}\label{eqn-1} 
    d_{CLIP}(X,Y) = 1 - \frac{e(X) \cdot e(Y)}{||e(X)||||e(Y)||}. 
\end{equation}
Where \mbox{$e(\cdot)$} is the image encoder of the CLIP model, and ClipScore is the cosine similarity between the image encodings.\par
~\\
\emph{D. Inspiration from Related Works}\par
In summary, pixel-based or structure-based metrics are limited by their computation process and have difficulty measuring semantic-level image differences. While existing semantic-level metrics also have their limitations.\par
Due to the limitations of the convolutional layer architecture, LPIPS still finds it difficult to truly understand semantic information from the perspective of the entire image. Nevertheless, this method has proven that using data-driven approaches to fit human perception is effective.\par
The same problem occurs in the ViTScore as well, although it introduces contextual and global information to a certain extent, this algorithm is still not robust enough to changes weakly correlated with image semantics such as translation or selection, as it is greatly affected by the patch segmentation positions. However, this is still a positive attempt. Although it uses neural networks, its structured processing of the neural network outputs greatly increases the interpretability of the algorithm. During the calculation, we can clearly see which patch in image \mbox{$X$} has higher semantic similarity with which patch in image \mbox{$Y$}. This decoupled combination of deep learning and traditional algorithms provides us with many insights.\par
ClipScore has advantages in measuring the loss of information at the semantic level, but it lacks interpretability and is limited by the bag-of-words algorithmic structure. It is known to easily confuse the subject and object relationships, such as mistaking "the horse eats the grass" for "the grass eats the horse". This could potentially introduce critical biases when computing the semantic similarity between images. These issues are important considerations that need to be carefully addressed when using ClipScore.\par
Based on the insights from the related works and their strengths and limitations, we hope our work can have the following characteristics:\par
\begin{enumerate}
    \item The metric should have structured and interpretable properties, with clear visual meanings.
    \item The metric should be optimized using human-annotated datasets, aligning better with human perception.
    \item The metric should not be limited to convolutional layers' restricted receptive fields, and should be able to fully understand the semantic information of the entire image.
    \item The metric should be able to "understand" the semantic content of the image, and clearly capture the objects and relationships between objects in the image, just like humans do.
    \item The metric should be robust to changes weakly correlated with image semantics, such as small translations or rotations.
\end{enumerate}   
In summary, the goal is to develop a more semantically-aware and human-aligned image similarity metric to better measure the differences of images transmitted by semantic communication systems.\par
\begin{figure*}[t]
    \centering
    \includegraphics[scale=1.2]{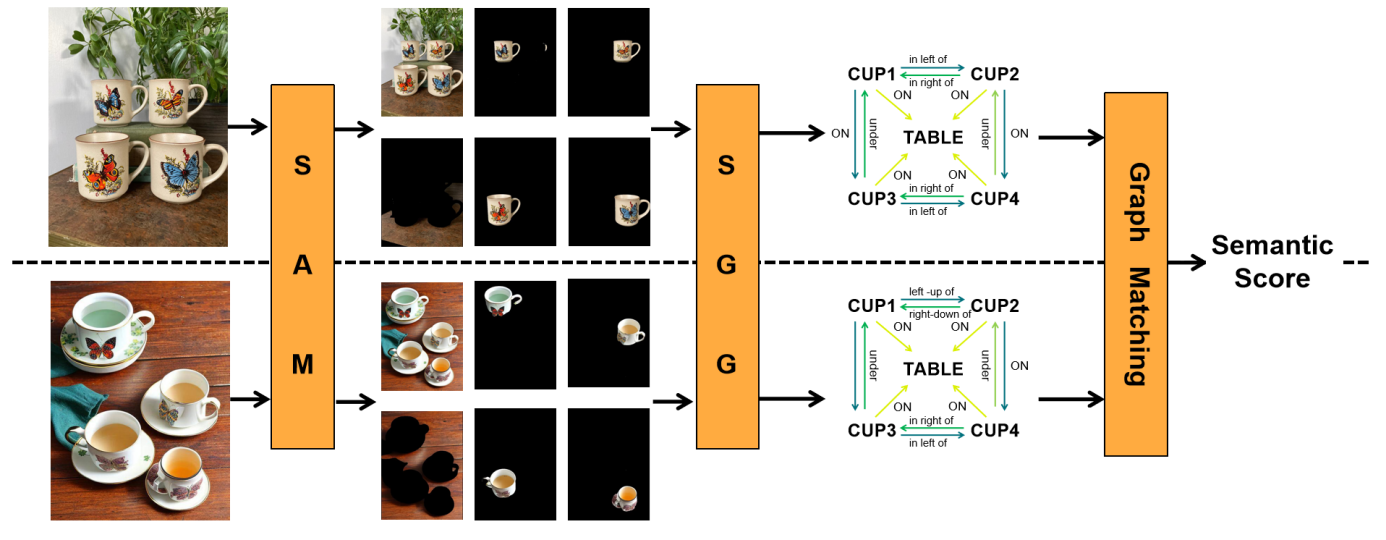}
    \caption{The architecture of SeSS. Images are segmented into object-level masks by model SAM\cite{sam}. Then SGG model get object-relation graphs based on images and masks. The matching score between graphs calculated by the graph matching algorithm shows the similarity score between images.}
\end{figure*}  

\section{Method}
Inspired by the related work in Section II, we decided to use the multimodal CLIP model as the basis, and reference the ViTScore approach to use structured algorithms to constrain its calculation, in order to increase the interpretability of the algorithm and avoid its problem of confusing the subject and object relationships. Scene Graph Generation technology and graph matching technology become good choices, with the former decomposing the image into a graph composed of objects and relationships between objects, and the latter performing structured similarity calculation on the graph. On this basis, referencing LPIPS, image semantic similarity data is manually annotated to tune the hyperparameters in the algorithm, so that its results are closer to human perception.\par
The overall architecture of SeSS is shown in Figure 2. The images first goes through the SAM segmentation model, which semantically segments the image into a set of object-level masks. This segmentation result is then input to the Scene Graph Generation model, which performs relational inference between the segmented objects, generating an object-relation graph. The object-relation graphs of the two images, along with the original images and the CLIP model, are then used for graph matching to compute the final semantic similarity score. This process aligns with the cognitive process of how humans perceive complex images.\par
Detailed introductions are provided from three aspects: (1) generating the object-relation graph using SAM+SGG, (2) the graph matching algorithm, and (3) hyperparameter tuning using human-annotated data.\par
~\\
\noindent \emph{A. Generating the Object-relation Graph using SAM+SGG}\par
When attempting to understand images at the semantic level, simply segmenting the different objects in the image is not enough. Models are also expected to be able to infer and predict the relationships between the various entities in the image. The task of Scene Graph Generation aims to solve this problem. In the paper\cite{psg}, the problem of Panoptic Scene Graph (PSG) generation was proposed. The PSG dataset introduced in this research contains nearly 50,000 COCO\cite{coco} images, with the relationships between the segmented objects in the COCO dataset annotated, and 56 common everyday relationships defined precisely.\par
This research proposed two types of predictive models, a two-stage approach and a single-stage approach. The two-stage approach decomposes the generation of the entire object-relation graph into two parts: (1) using a pre-trained Panoptic FPN\cite{panoptic} model for instance segmentation to extract initial object features, masks, and class predictions, and then (2) using classic Scene Graph Generation methods (such as IMP\cite{imp}, MOTIFS\cite{motifs}, VCTree\cite{vctree} and GPSNet\cite{gpsnet}) to process this and obtain the final scene graph prediction. The single-stage approach, on the other hand, combines the two components into one end-to-end trainable model.\par
Considering the limitations of the single-stage prediction model due to the object label constraints of the COCO dataset, the algorithm adopts a two-stage prediction architecture, and also replaces the panoptic segmentation model with the SAM segmentation model. SAM, as a large-scale image segmentation model, is trained on a dataset containing 11 million images and over 1 billion segmentation masks. It is able to segment all objects in an image, no longer constrained by the object labels in the COCO dataset. This significantly improves the overall versatility and generalization capability of the algorithm.\par
As shown in Figure 3,  the input image \mbox{$I$} goes through the SAM model, which outputs the image masks \mbox{$M$} and their corresponding confidence scores.\par
\begin{equation}\label{eqn-1} 
    Pr(M|I) = SAM(I).
\end{equation}
The PSG model takes the masks and confidence scores as input, and computes the scene graph \mbox{$G$},\par
\begin{equation}\label{eqn-1} 
    Pr(G|I) = Pr(M|I)Pr(R|M, I).
\end{equation}
Where \mbox{$I$} is the input image, \mbox{$G$} is the scene graph, i.e. the desired object-relation graph, which includes the mask images \mbox{$M = \{m_1, ..., m_n\}$} of \mbox{$n$} objects in the image, as well as the set of relationships \mbox{$R \in \{r_1, ..., r_l\}$} between them. Specifically, \mbox{$m_i$} represents the mask matrix of object \mbox{$i$}, and \mbox{$r_i$} belongs to the set of all relationship classes.\par

\begin{figure*}[t]
    \centering
    \includegraphics[scale=0.34]{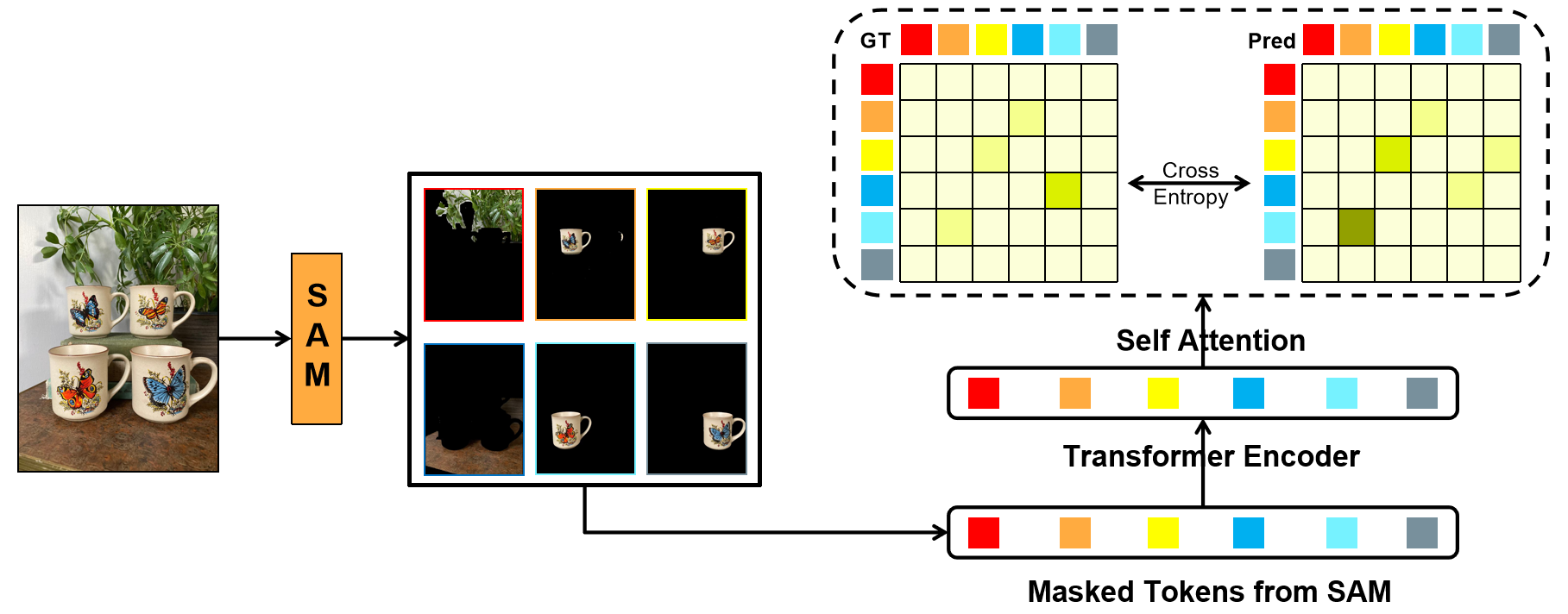}
    \caption{The process of converting images into object-relation graphs based on model SAM and PSG. The PSG model using transformer encoder and self-attention technology to get predict relation matrix from masked tokens given by SAM. Using cross entropy between the ground truth matrix and the predict one, PSG model learns to predict relationshipe of objects masked by SAM.}
\end{figure*}  
\begin{figure*}[t]
    \centering
    \includegraphics[scale=0.55]{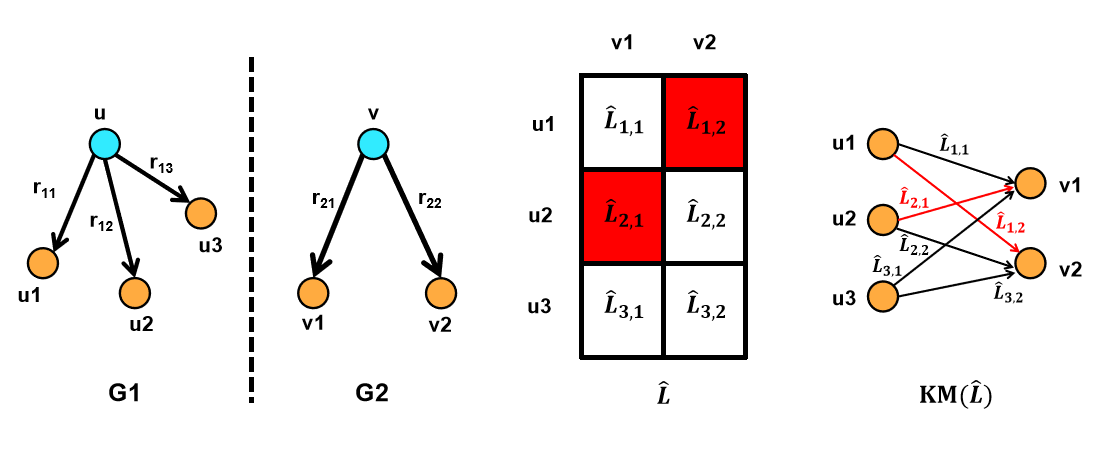}
    \caption{Visual example of calculating node similarity between \mbox{$u$} in \mbox{$G_1$} and \mbox{$v$} in \mbox{$G_2$}. The node similarity is calculated by the similarities of their neighboring nodes, and their relevant relation lables. The problem is transformed into a bipartite graph maximum matching problem in matrix form, which is solved using the KM algorithm to obtain the similarity score.}
\end{figure*}  

~\\
\emph{B. Graph Matching Algorithm}\par
After processing the image through SAM+SGG, the corresponding object-relation graph is obtained. The problem of calculating the semantic similarity between images is then transformed into the problem of matching the object-relation graphs corresponding to the images. This paper designs an iterative graph matching algorithm to calculate the matching score between object-relation graphs, which is the final semantic similarity metric between the images.\par
First, the overall ClipScore between the two images is calculated,\par
\begin{equation}\label{eqn-1} 
    image\_score = ClipScore(I_1, I_2).
\end{equation}
This \mbox{$image\_score$} is used as part of the semantic similarity metric. It is then combined with the graph matching score through a weighted sum to get the final score. The introduction of this \mbox{$image\_score$} serves two purposes. It incorporates the overall visual difference between the images, and it also allows the algorithm to handle images that do not contain objects, such as noise patterns.\par
Then all the nodes in the scene graph \mbox{$G_1$} corresponding to image \mbox{$I_1$} and the scene graph \mbox{$G_2$} corresponding to image \mbox{$I_2$} are used to calculate their initial similarity scores in a pairwise manner,\par
\begin{equation}\label{eqn-1} 
    L_{i,j} = ClipScore(I_1[mask(u_i)], I_2[mask(v_j)]).
\end{equation}
\mbox{$u$} and \mbox{$v$} represent the nodes in \mbox{$G_1$} and \mbox{$G_2$} respectively. The initial similarity score between node \mbox{$u_i$} in \mbox{$G_1$} and node \mbox{$v_i$} in \mbox{$G_2$} is their ClipScore between the corresponding mask images. This gives the initial similarity matrix \mbox{$L$}. The ClipScore is able to better capture the semantic-level information in the mask images. Additionally, since the objects are segmented individually, there is no issue of confusing the main and secondary subjects. Therefore, the ClipScore is used as the basis for calculating the similarity between the objects.
Solely calculating the visual similarity between objects is not enough. The different relationships that the object nodes are in, as well as the connected nodes, i.e. the topological position of the object within the overall relation graph, also determines the similarity between the objects. Therefore, an iterative approach is used, where the similarity between neighboring nodes is propagated, and the entire similarity matrix is continuously updated. This allows for the most reasonable object matching results to be obtained. The iterative method is used to update the similarity matrix.\par
Considering the similarity calculation between node \mbox{$u$} in graph \mbox{$G_1$} and node \mbox{$v$} in graph \mbox{$G_2$} during each update iteration, the visual example is shown in the Figure 4, where node \mbox{$u$} has neighboring nodes \mbox{$u_1$}, \mbox{$u_2$}, \mbox{$u_3$}, and node \mbox{$v$} has neighboring nodes \mbox{$v_1$}, \mbox{$v_2$}. \par
The similarity between \mbox{$u$} and \mbox{$v$} is determined by the similarity \mbox{$L$} obtained from the previous iteration, as well as the matching degree of their neighboring nodes. The neighboring node matching is a small-scale graph matching problem, resulting in a 3x2 matrix \mbox{$\hat{L}$}, \mbox{$\hat{L}_{k,l}$} representing the matching degree between \mbox{$u_k$} and \mbox{$v_l$}. The calculating process is as follows,\par
\begin{equation}\label{eqn-1} 
    \hat{L}_{k,l} = \alpha L_{k,l} + (1-\alpha) R_{r_{1k}, r_{2l}}.
\end{equation}
Where \mbox{$\alpha$} is the weight between node similarity and relation similarity. The similarity between \mbox{$u_k$} and \mbox{$v_l$} in this subgraph is obtained by a weighted sum of the similarity \mbox{$L_{k,l}$} computed in the previous iteration, and the similarity of their relations relative to nodes \mbox{$u$} and \mbox{$v$}. The relation similarity is calculated based on the textual similarity between the relations.
\begin{equation}\label{eqn-1} 
    R_{i,j} = ClipScore(r_i, r_j).
\end{equation}

After all the calculations of \mbox{$\hat{L}$} are complete, the maximum matching problem is transformed into a maximum sum problem of selecting non-overlapping rows and columns from a non-negative matrix. This is a classic bipartite graph maximum matching problem. By constructing a bipartite graph based on the matrix, the maximum matching value between the neighboring nodes of \mbox{$u$} and \mbox{$v$} can be obtained using the KM algorithm\cite{km}. This value is then used to update \mbox{$L_{u,v}$},\par
\begin{equation}\label{eqn-1} 
    L_{u,v}^{'} = (1 - \beta ) L_{u,v} + \beta KM( \hat{L} ).
\end{equation}
Parameter \mbox{$ \beta $} is the update rate of the iterative algorithm. After all the node pairs between \mbox{$G_1$} and \mbox{$G_2$} have been processed and the iterated matrix \mbox{$L^{'}$} is obtained, the values in \mbox{$L^{'}$} are used to update the matrix \mbox{$L$}.\par
After all the iterations are complete, the maximum matching value in the final similarity matrix \mbox{$L$} will be used as a component of the final SeSS,\par
\begin{equation}\label{eqn-1} 
    SeSS = (1- \gamma )*KM(L) + \gamma*image \_ score.
\end{equation}
In simple terms, the final SeSS is obtained by matching each object in image 1 with an object in image 2, and using the max sum of the matching values between the nodes as the semantic similarity metric between the two images.\par
It's worth noting that when actually computing the KM maximum matching value, the algorithm performs a weighted processing based on the semantic importance of the nodes in the corresponding Graph. Based on the image and the segmented object masks, the semantic importance of different objects in the image is calculated, which is then applied in the maximum matching value calculation.\par
\begin{equation}\label{eqn-1} 
    I \_ imp = imp \_ predict(I).
\end{equation}

\begin{equation}\label{eqn-1} 
    obj \_ imp(o_i) = \sum I \_ imp [mask(o_i)].
\end{equation}
Where \mbox{$imp \_ predict(\cdot)$} is the image pixel importance prediction algorithm. It predicts the visual importance of different pixels in the image based on the visual structure of the image and the rate of change of pixel values. \mbox{$mask(o_i)$} is the mask of object \mbox{$o_i$}, and the semantic importance of object \mbox{$o_i$} is the sum of the pixel importance values of the pixels covered by its mask. Then the semantic importance of all objects is normalized into a probability distribution array \mbox{$ \{ imp_1, imp_2, ..., imp_n \} $} that sums to 1.\par
The iterative process of this graph matching algorithm progressively increases the final matching values for node pairs that are similar at the semantic relationship hierarchy level, while decreasing the matching values for node pairs that are quite different at the semantic relationship hierarchy level. For example, an image of a person eating and an image of a cow eating grass would be assigned an extremely low similarity score by traditional image metrics. However, these two images do exhibit semantic similarity, as the subject is engaged in the act of eating the object in both cases. In the iterative algorithm, the subject and object entities are ultimately matched through the common "eating" relationship between them, even though the final matching value may not be very high. This approach aligns with human cognitive patterns and can better capture the differences at the semantic hierarchy level.\par

~\\
\emph{C. Hyperparameter Tuning using Human-annotated Data}\par
According to the LPIPS metric, it is believed that optimizing the algorithm using human-annotated image similarity score data can enhance its performance and bring it closer to human perception. We have manually annotated the semantic similarity scores for a hundred thousand sets of similar images, as shown in the figure 5. Each data set contains three images - the original image and two images that have a certain degree of similarity to the original. These images were either collected from open-source datasets or images generated using large-scale image generation models.\par
These annotated image pairs are used to optimize the hyperparameters in the graph matching algorithm, such as the \mbox{$\alpha$}, \mbox{$\beta$}, and \mbox{$\gamma$} parameters shown in the subsection B formula. The number of iterations and other hyperparameters are also optimized in this process. Specifically, the optimization process uses hyperparameter search\cite{random}, constraining the weighted sum hyperparameters like \mbox{$\alpha$}, \mbox{$\beta$}, and \mbox{$\gamma$} constrained to the range of 0-1. This provides a search space for tuning the hyperparameters. The optimal hyperparameters are reported in Table II.\par
\begin{figure}[h]
    \centering
    \includegraphics[scale=0.32]{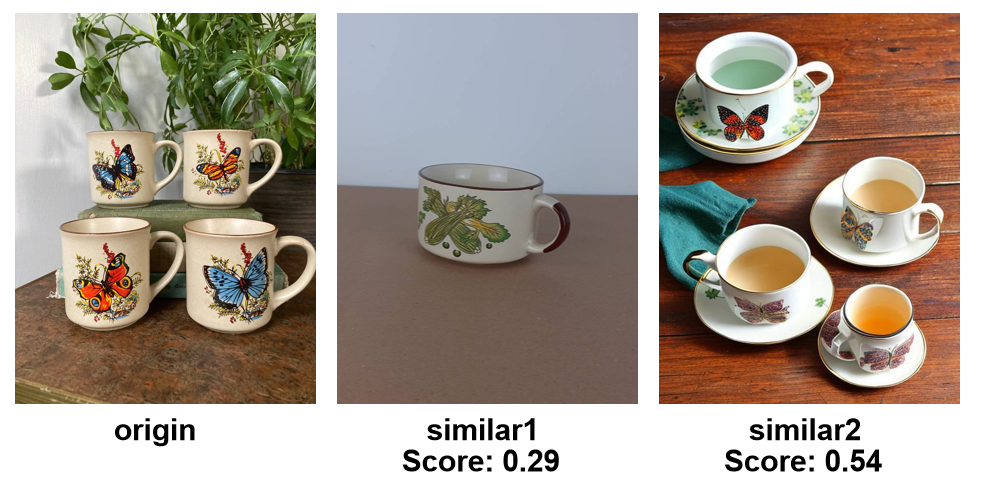}
    \caption{Visual example of human-annotated similarity scores of image pairs. 
    The leftmost image is the original image, and the two images to the right are similar images to the original. Human annotations have been provided to respectively indicate the similarity between the two similar images and the original image.}
\end{figure}  
\begin{table}[ht]
    \renewcommand\arraystretch{1.7}
    \begin{flushleft}   
    \caption{Hyperparameter of the Graph Matching Algorithm} 
	\centering
    \label{table:1} 
	\begin{tabular}{|m{1.2cm}|m{1.2cm}|m{5.0cm}|} 
        \hline Hyper-parameter & Optimal Values & Meanings \\ 
        \hline $\alpha$ & 0.25 & The weighting between importance of object and relationship when computing neighboring node similarity \\
        \hline $\beta$ & 0.05 & The rate of convergence of the object similarity matrix iteration \\
        \hline $\gamma$ & 0.10 & The contribution of the ClipScore between images to the overall metric score \\ 
        \hline ite & 7 & The number of iterations for computing the object similarity matrix \\
        \hline k & 2.25 & The square root of the normalization factor for the object semantic importance weighting \\
        \hline 
	\end{tabular}
\end{flushleft} 
\end{table}
\begin{figure}[ht]
    \centering
    \includegraphics[scale=0.8]{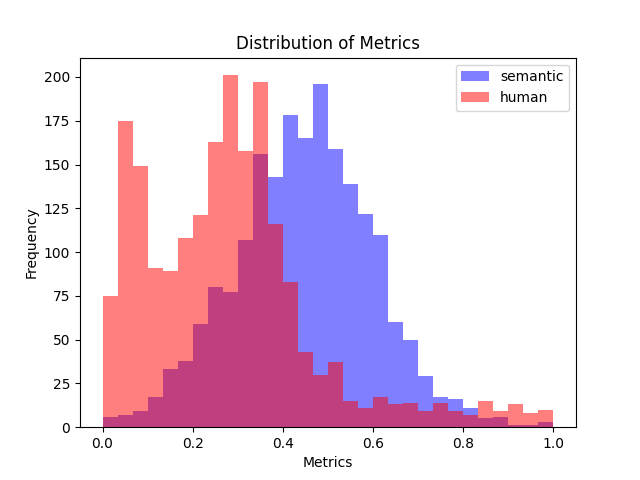}
    \caption{Distribution of the human-annotated similarity score(red) and score calculated by SeSS(blue).}
\end{figure}  

The optimization results in table II provide some interesting findings. For example, humans tend to care more about the degree of object matching itself rather than the relationships between objects when making semantic-level similarity comparisons. For instance, the semantic similarity between "person eating" and "person cooking" tends to be higher than the similarity between "person eating" and "horse eating grass". These findings were used to help fine-tune the SeSS to better align with human perception of the semantic information in images.\par
As shown in Figure 6, the distribution of SeSS after hyperparameter tuning of the graph matching algorithm (in blue) and the distribution of human-annotated scores (in red) exhibit some differences. Since the human raters applied more stringent standards in their manual scoring, they tended to assign lower semantic similarity scores compared to the structured SeSS. On average, the human-annotated scores are slightly lower than the algorithmic scores by less than 0.2.\par

\section{Experiment Results}
This section is mainly introduced the relevant testing settings, including the dataset for SeSS's test, the introduction of baseline and the performance for the SeSS in different datasets.\par
Discription and figures are given to show how the SeSS surpass the existing image similarity metrics when measuring the difference between images at semantic level.\par
~\\
\emph{A. Experiment Description}\par
The SeSS aims to measure the information loss of visual communication systems during the transmission process at the semantic level. It’s hoped that the semantic-level metric can work normally in both traditional and semantic communication systems, and can serve as a bridge for comparing the two, while maintaining sufficient robustness to changes weakly correlated with image semantics. In order to verify these characteristics, four experiments based on different data have been conducted, including: (1) The image set transmitted by the traditional and semantic communication image systems under different compression ratios. (2) The image set transmitted by the traditional and semantic communication image systems under different signal-to-noise ratios. (3) The set of similar images generated by large models with different noises introduced. (4) Image cases after certain special transformations.\par
The experimental data is sampled from the COCO2017 dataset. In order to fully demonstrate the performance of the semantic-level metric, six representative metrics, including PSNR, SSIM, MS-SSIM, LPIPS, ViTScore, and ClipScore, are used for comparison. At the same time, many visual examples will be presented to better illustrate the effectiveness and interpretability of the SeSS.\par
~\\

\begin{figure*}[ht]
    \centering
    \includegraphics[scale=0.35]{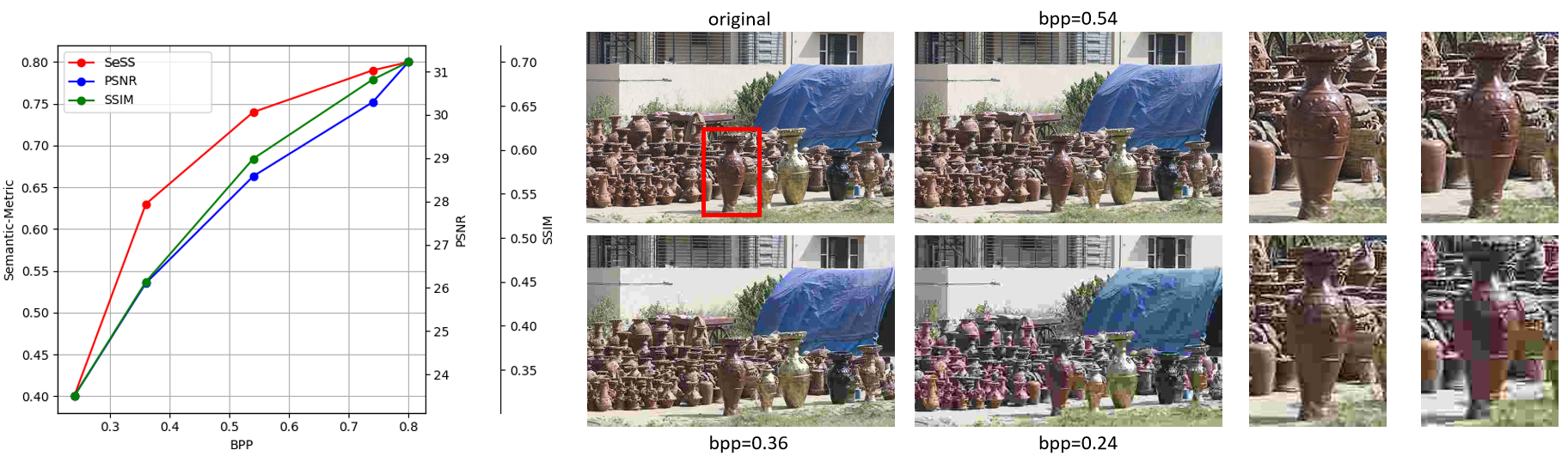}
    \caption{The similarity score of SeSS, PSNR and SSIM on the JPEG2000 dataset under different bpp (bits per pixel) values, where the red line represents SeSS. The right-hand side shows the specific visual examples under different JPEG2000 compression bpp values, which reveals that as the bpp decreases to a certain extent, the objects in the image become difficult to recognize, and human understanding of the semantic-level information in the image will suffer significant degradation. This trend is consistent with the behavior of the SeSS metric.}
\end{figure*}  
\begin{figure*}[ht]
    \centering
    \includegraphics[scale=0.35]{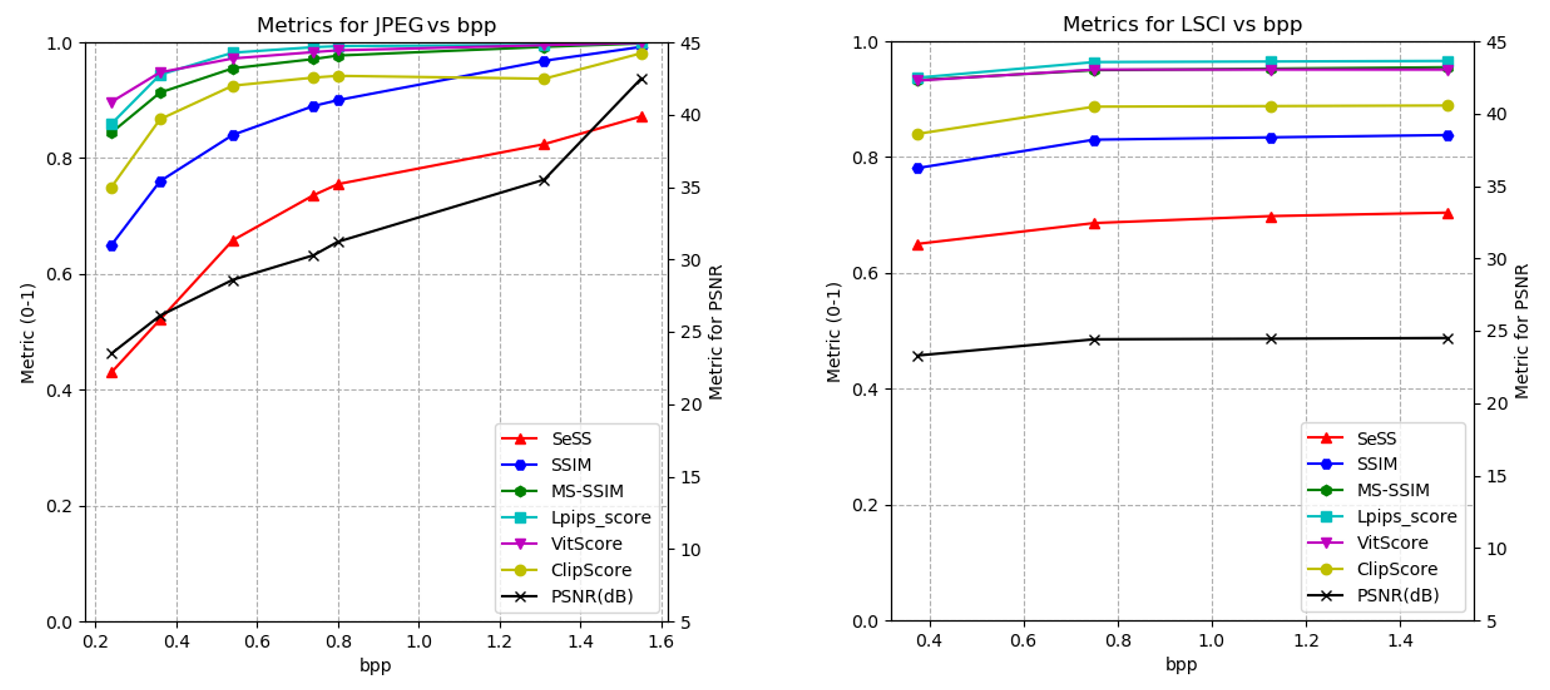}
    \caption{The similarity score of SeSS, PSNR, SSIM, MS-SSIM, LPIPS, ViTScore and ClipScore on the images compresed by JPEG and LSCI under different bpp values, where the red line represents SeSS. Each subfigure can demonstrate the similarities and differences in trends and numerical values between the SeSS metric and other metrics.}
\end{figure*}

\noindent \emph{B. Metric Variations under Different Compression Ratios}\par
Figure 7 compares the SeSS, PSNR, and SSIM metrics of JPEG2000 at different bpp (bits per pixel) levels. It can be observed that as the compression ratio increases and the image details are continuously lost, the SeSS score also decreases accordingly, indicating that it has the capability to evaluate traditional image compression algorithms. In the horizontal comparison, the trends show that PSNR and SSIM are relatively even with respect to changes in bpp, but the changes in SeSS are more gradual in the high bpp region and decline rapidly in the low bpp region. This is consistent with human perception of images, as shown in the top right subfigure. When the compression rate is relatively low, the slight blurring and distortion in the image have a relatively small impact on the human's semantic-level perception of the image. However, once a certain threshold is reached, the blurring and distortion make it difficult to recognize the content, and the human's semantic-level perception of the image will decline rapidly.\par
Figure 8 shows the metric values of the JPEG image compression algorithm and the LSCI image semantic communication system\cite{lsci} at different bpp (bits per pixel) levels, respectively. In each subfigure, the SeSS decreases as the bpp decreases, both in the traditional image compression algorithm and the deep learning-based semantic communication algorithm. This indicates that the SeSS metric can serve as a foundation for evaluating the performance of these algorithms. In contrast, other metrics such as MS-SSIM, LpipsScore, ViTScore, and ClipScore do not show obvious perception of image information loss at high bpp levels. In fact, metrics like ClipScore even exhibit the abnormal phenomenon of higher scores at lower bpp levels. \par
\begin{figure*}[ht]
    \centering
    \includegraphics[scale=0.4]{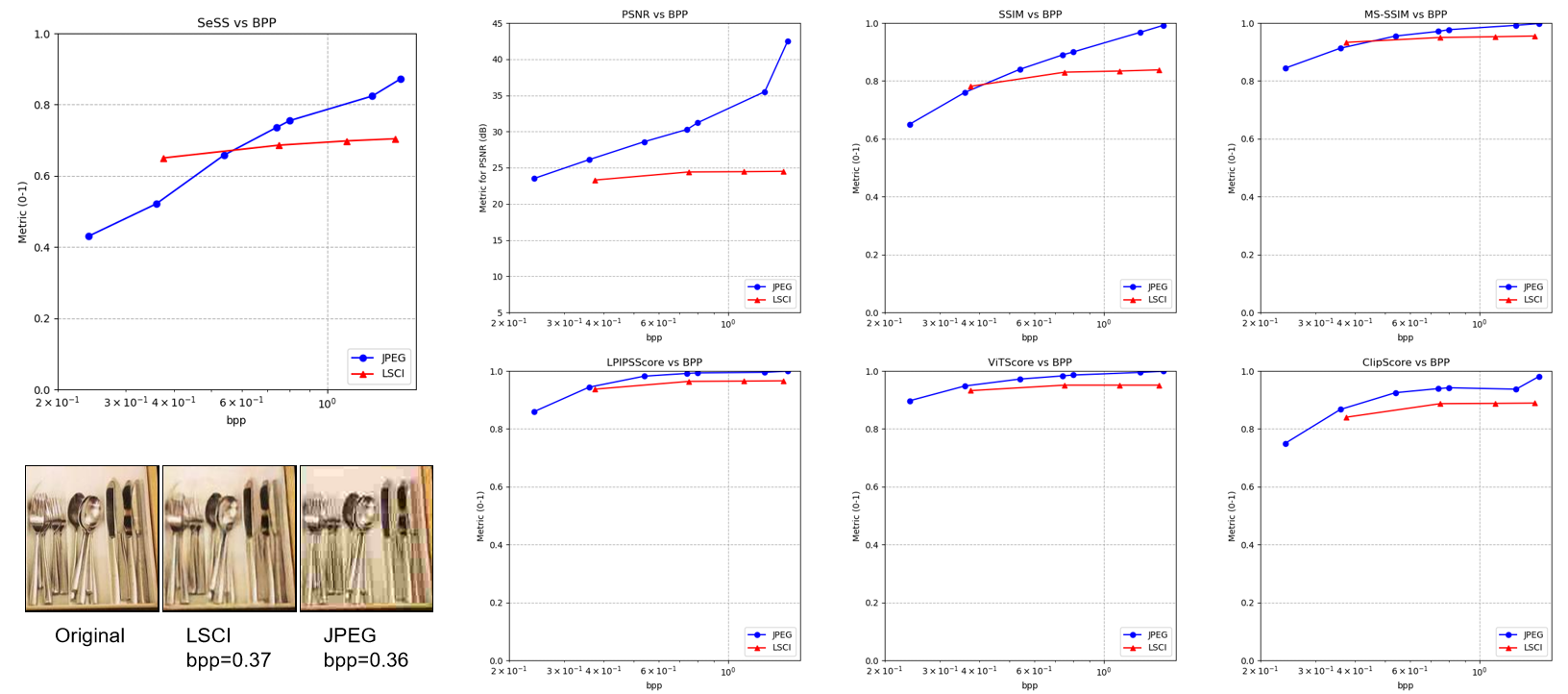}
    \caption{Horizontal comparison between the two systems in metric of SeSS, PSNR, SSIM, MS-SSIM, LPIPS, ViTScore and ClipScore under different bpps. To better showcase the experimental results at low bpp, the x-axis of the figures is presented in logarithmic scale. At lower bpps, LSCI prioritizes the preservation of semantic-level information. SeSS gives LSCI a higher score at low bpps, which can help to show the advantage of the semantic communication system in this aspect. The visual examples are provided in the subfigure in the bottom left.}
\end{figure*} 
\begin{figure*}[ht]
    \centering
    \includegraphics[scale=0.35]{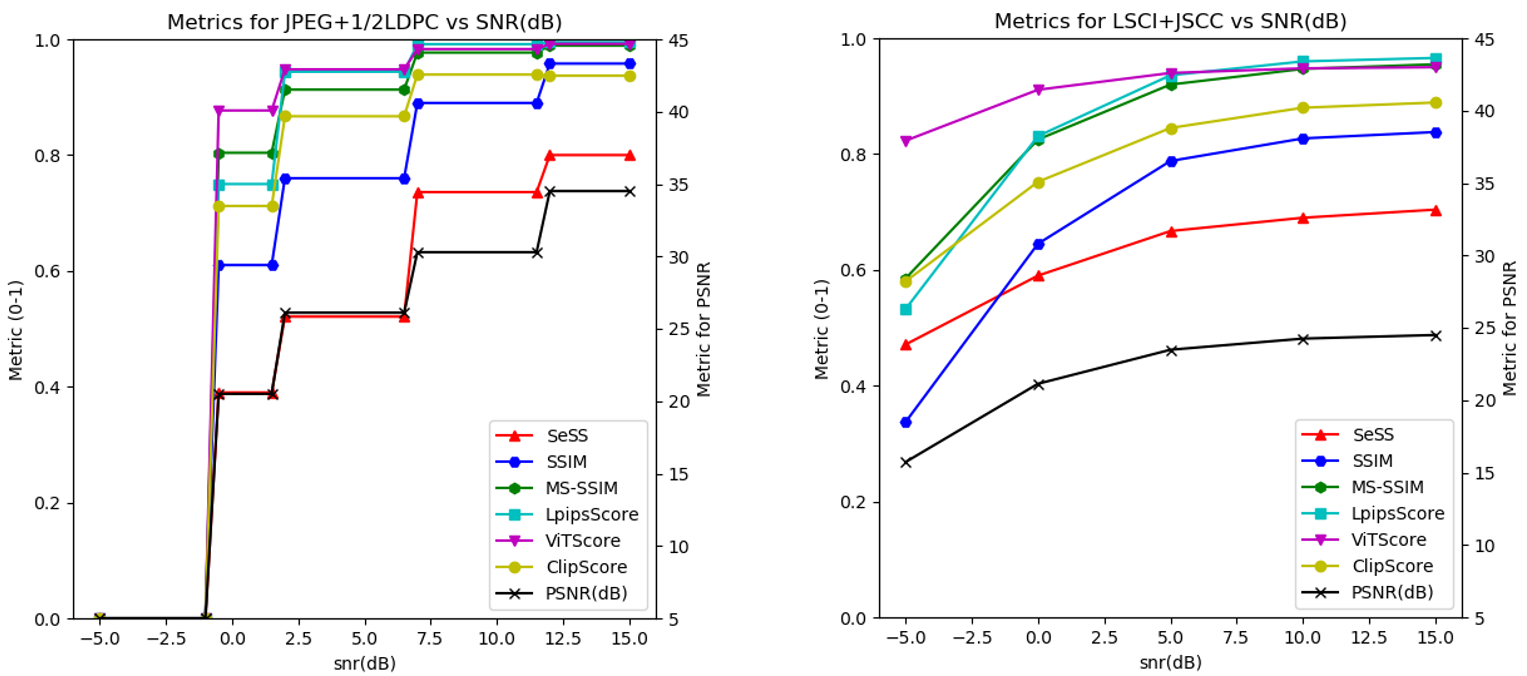}
    \caption{The similarity score of SeSS, PSNR, SSIM, MS-SSIM, LPIPS, ViTScore and ClipScore on the images transferred by traditional communication pattern JPEG2000+1/2LDPC and semantic communication system LSCI+JSCC under AWGN channel with different SNRs, where the red line represents SeSS. Each subfigure can demonstrate the similarities and differences in trends and numerical values between the SeSS metric and other metrics. The LSCI system uses 16QAM modulation. The JPEG+1/2LDPC system uses BPSK, QPSK, 16QAM, and 64QAM modulation schemes sequentially at different signal-to-noise ratios. The experiment ensures that the system maintains a constant \mbox{$ \frac{Image \; Compression \; Rate}{Bits \; per \; Symbol} $} during transmission.}
\end{figure*} 
Figure 9 provides a horizontal comparison between the two systems. It can be observed that under the evaluation of the SeSS, the semantic communication system outperforms the traditional system at lower compression rates, which is consistent with the actual visual perception. This is because the LSCI semantic communication architecture based on an end-to-end design prioritizes the preservation of important semantic information in the image at higher compression ratios.\par
In contrast, other metrics are unable to reflect this advantage of the semantic communication system. In fact, there can be a significant gap in the metric values between the two systems, making it impossible to place traditional communication systems and semantic-based image communication systems on the same metric scale for performance comparison.\par

~\\
\emph{C. Metric Variations under Different Signal-Noise-Ratio}\par
Subsection B proves that the SeSS can evaluate different source coding and decoding systems across various compression rates. Furthermore, since the SeSS is designed to assess visual semantic communication systems, it is necessary to demonstrate its ability to measure the similarity of transmitted images under different channel conditions.\par
The subfigures in figure 10 respectively illustrate the performance of the traditional communication pattern JPEG+1/2LDPC and semantic communication system LSCI using joint source-channel coding (JSCC) technique\cite{jscc, jscc-f} under simulated AWGN channel with different SNRs. The LSCI system uses 16QAM modulation. The JPEG+1/2LDPC system uses BPSK, QPSK, 16QAM, and 64QAM modulation schemes sequentially at different signal-to-noise ratios. The experiment ensures that the system maintains a constant \mbox{$ \frac{Image \; Compression \; Rate}{Bits \; per \; Symbol} $} during transmission. It can be observed that the metric behaves normally under both LDPC coding and JSCC, with the metric scores decreasing as the signal-to-noise ratio (SNR) of the channel decreases.\par
Figure 11 makes a horizontal comparison between the two systems under different SNRs. LSCI using JSCC technique to fight against noises, which prioritizes the preservation of semantic-level information under lower SNRs. SeSS gives LSCI a higher score at low SNRs, which can help to show the advantage of the semantic communication system in this aspect. The visual examples show LSCI+JSCC under 5dB SNR is more similar to the original image at semantic level as well as the human perception.\par
\begin{figure*}[ht]
    \centering
    \includegraphics[scale=0.4]{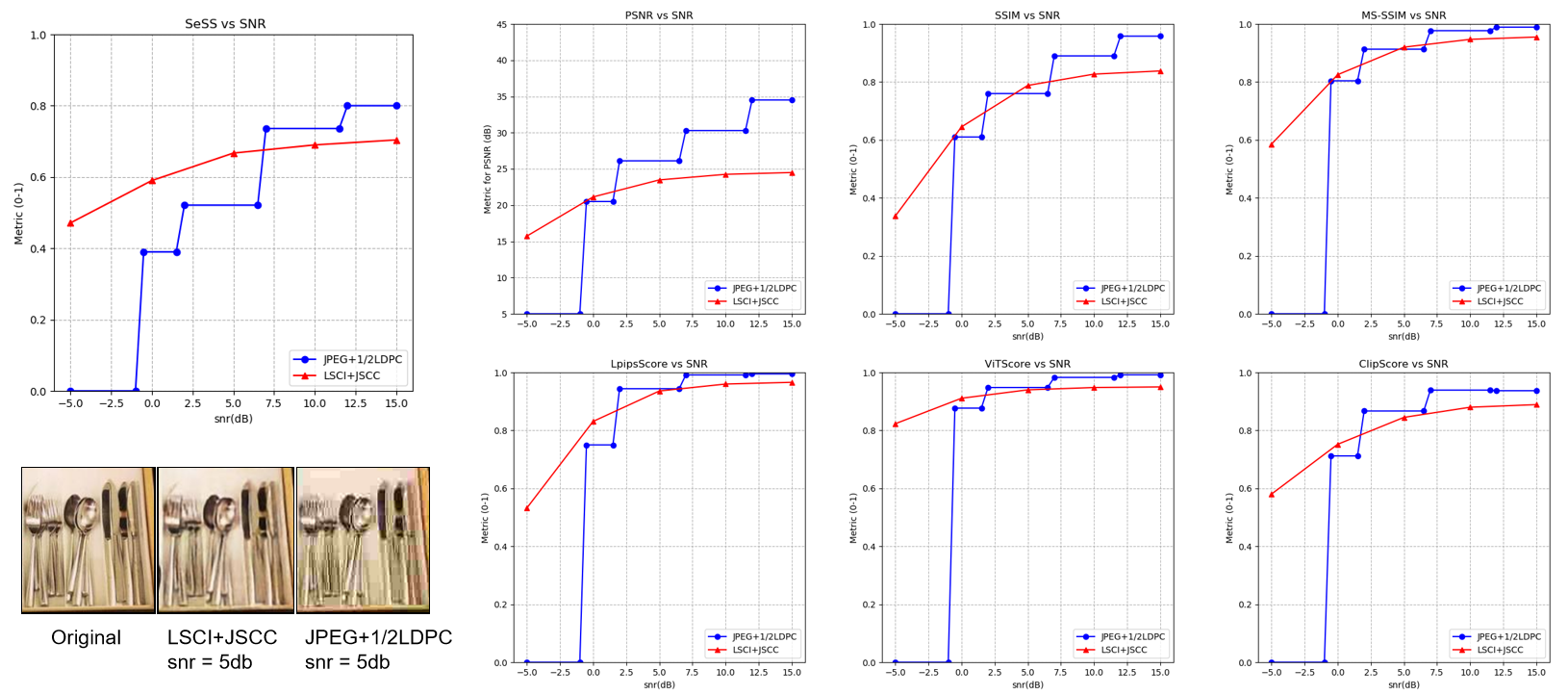}
    \caption{Horizontal comparison between the two systems in metric of SeSS, PSNR, SSIM, MS-SSIM, LPIPS, ViTScore and ClipScore under different SNRs. The visual examples are provided in the subfigure in the bottom left.}
\end{figure*} 
\begin{figure}[h]
    \centering
    \includegraphics[scale=0.5]{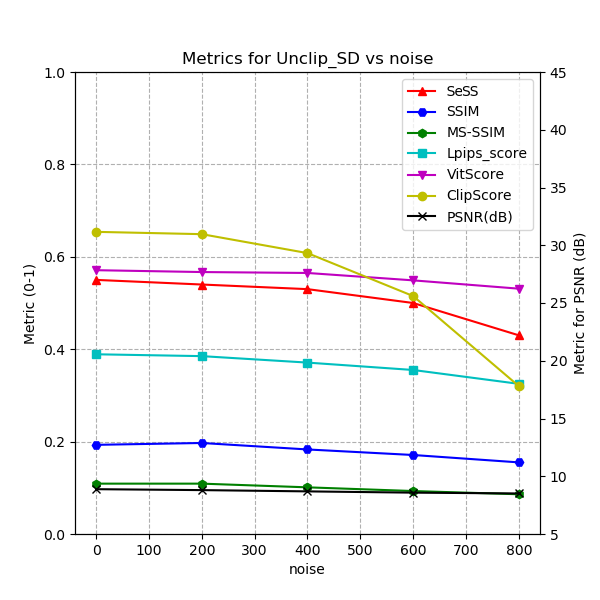}
    \caption{The similarity score of SeSS, PSNR, SSIM, MS-SSIM, LPIPS, ViTScore and ClipScore on the images generated by model Unclip with different noise. As the noise level increases, the generated images become less similar to the original images, both in terms of semantic level and visual aspects.}
\end{figure} 
\begin{figure*}[ht]
    \centering
    \includegraphics[scale=0.38]{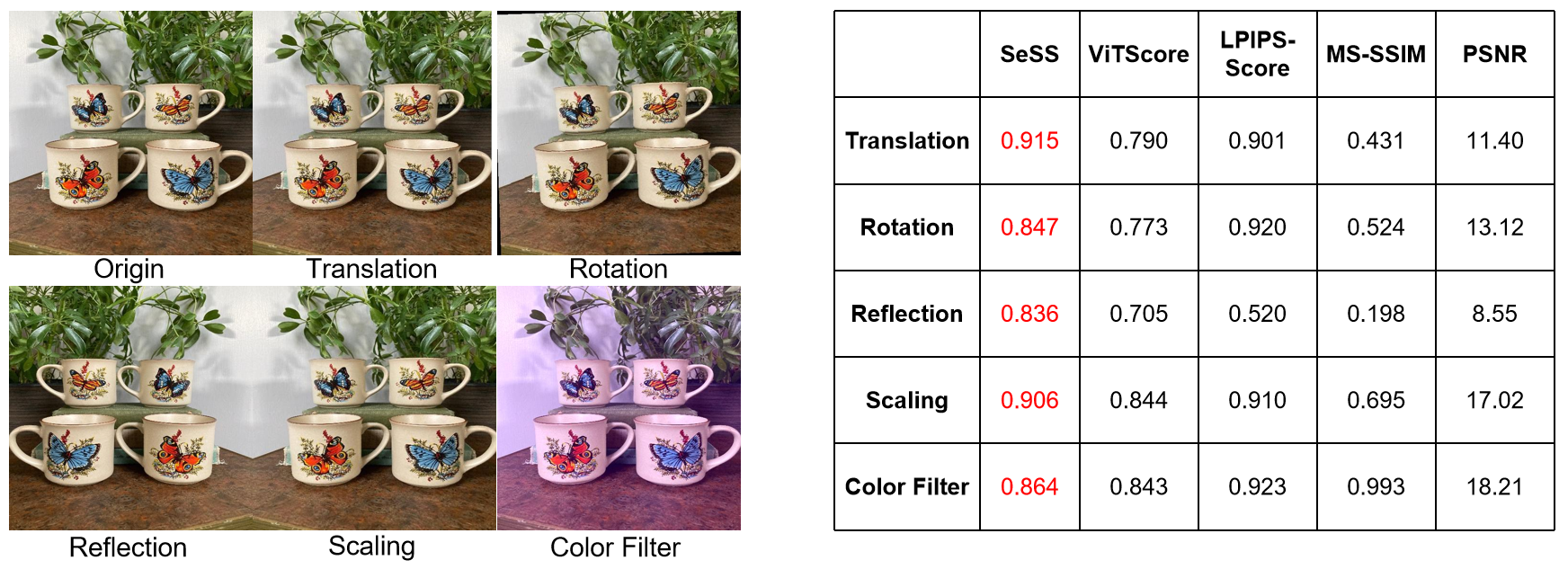}
    \caption{Visual examples and experiment statistic to illustrate the robustness of different metrics under changes weakly correlated with image semantics. Changes list here include small image translations, rotations, reflections, scaling, and color filter changes. SeSS is the most robust metric against these transformations.}
\end{figure*} 
~\\ 
\emph{D. Metric Variations under Different Generation Noise}\par
Figure 12 shows the performance of the Unclip\cite{unclip} model-generated similar images on various metrics. The Unclip model introduces noise to control the semantic-level similarity between the generated images and the original images.\par
It can be observed that the semantic-level metrics are more tolerant in scoring the similar images, whereas the traditional metrics give extremely low scores, rendering them incapable of measuring the semantic similarity between the generated images and the original images. Among the semantic-level metrics, LPIPS and ViTScore, which are based on image patches, are less sensitive to the semantic-level information changes caused by the introduced noise.\par
In contrast, ClipScore and SeSS are better able to measure the changes in the semantic-level information of the generated images, relative to the original images, under the influence of different noise strengths introduced by the generative model.\par
~\\
\emph{E. Image Cases after Certain Special Transformations}\par
Figure 13 uses visual examples to illustrate the robustness of different metrics under changes weakly correlated with image semantics, which list here include small image translations, rotations, reflections, scaling, and color filter changes. From a visual perspective, these changes have little impact on the semantic-level information in the images, and have a relatively small effect on human recognition of the semantic information. However, traditional pixel-level or structure-level metrics like PSNR or MS--SSIM give extremely low similarity scores, and patch-based metrics like LpipsScore and ViTScore also exhibit large fluctuations. The SeSS metric, on the other hand, demonstrates good robustness, indicating that it measures the difference in semantic-level information between images more effectively, making it more suitable for evaluating semantic-related tasks.\par
\begin{figure}[h]
    \centering
    \includegraphics[scale=0.65]{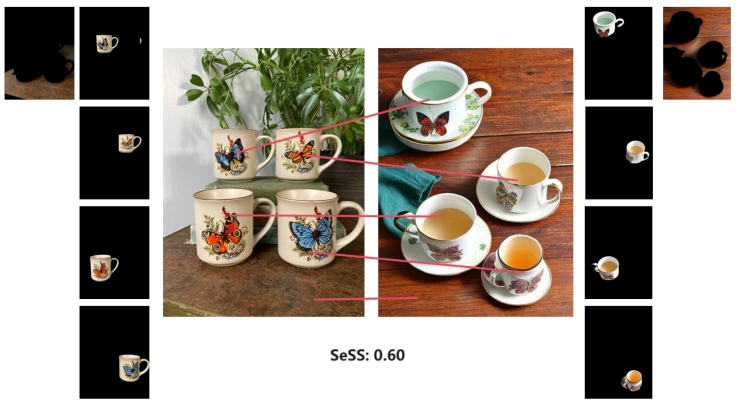}
    \caption{Specific visual example illustrates the matching results of the SeSS metric during the calculation process. From a semantic perspective, the matching result closely aligns with human perception.}
\end{figure} 
Figure 14 showcases the metric calculations and visual examples, demonstrating that the SeSS is able to structurally understand the semantic relationships within images, in alignment with the human process of comprehending the semantic content of images. The metric can comprehensively consider the visual content and semantic expression of images to evaluate the preservation and degradation of semantic information during the transmission process. Furthermore, the metric exhibits desirable properties such as being structured and interpretable.\par

\section{Conclusion}
In this paper, a novel image similarity metric SeSS is proposed, which is based on the SAM segmentation model, Scene Graph Generation technique, and object-relation graph matching approach. The computation process of this metric exhibits structural and interpretable properties, enabling it to measure the semantic-level differences between images. The similarity of image pairs has been manually annotated to optimize the hyperparameters of the graph matching algorithm within the SeSS, making the metric better aligned with human semantic perception. Subsequently, experiments were conducted to validate the performance of the SeSS.\par
However, SeSS still has limitations. The SeSS values are often confined to the range above 0.3, failing to fully utilize the 0-1 interval. Furthermore, due to the internal algorithm architecture, the metric tends to perform better in evaluating the similarity of complex images containing multiple objects, while its accuracy may be inevitably lower for images without objects.\par

\appendix
\begin{figure*}[ht]
    \centering
    \includegraphics[scale=0.42]{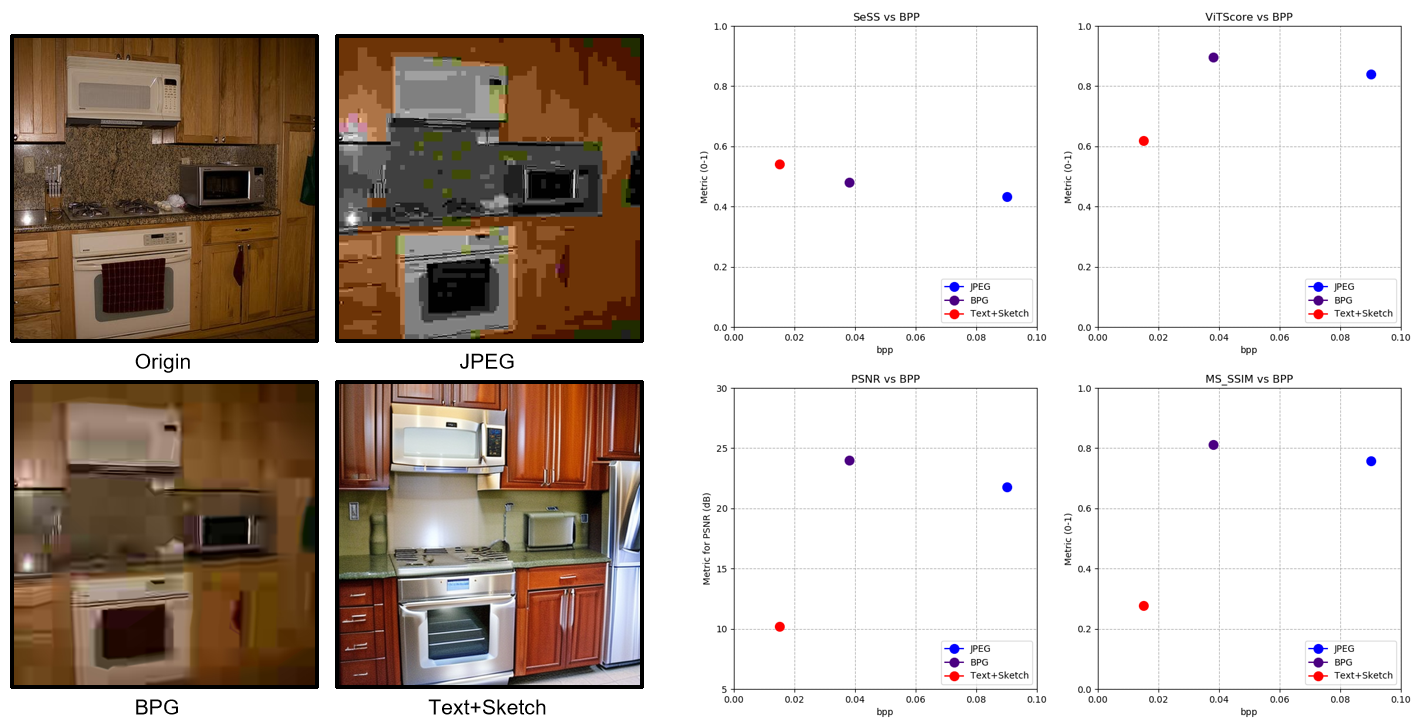}
    \caption{The similarity score of SeSS, ViTScore, PSNR, and MS-SSIM on the images compressed by JPEG, BPG, and the algorithm Text+Sketch at extreme low bpps. JPEG and BPG both reach their highest compression rates, with bpp at 0.090 and 0.038 respectively. The algorithm Text+Sketch compresses the images by extracting their natural language descriptions and Sketch features, and then generates the images using the Stable Diffusion model controlled by Controlnet, achieving a bpp of 0.015. According to the visual examples on the left, at extremely low compression rates, the JPEG and BPG compression algorithms exhibit significant distortion and blurriness, making the semantic information in the images difficult to recognize. In contrast, although Text+Sketch has differences in color and texture compared to the original image, it retains the main semantic information of the original image, and therefore receives a relatively higher SeSS score. However, on other metrics, the scores of the Text+Sketch algorithm are significantly lower than those of traditional compression algorithms.}
\end{figure*} 

\begin{figure*}[ht]
    \centering
    \includegraphics[scale=0.35]{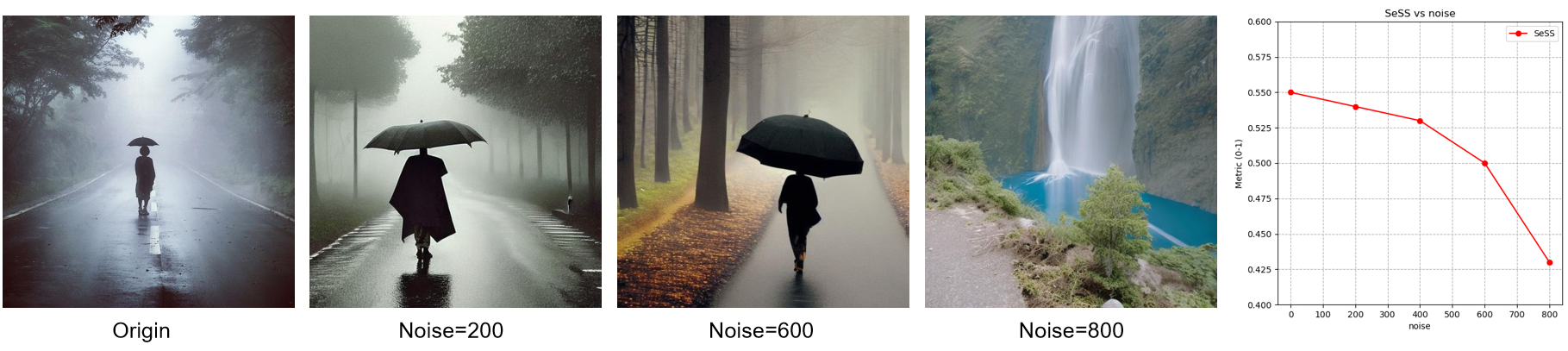}
    \caption{Supplementary explanation for Experiment C. The Unclip model introduces different levels of noise to output images with varying degrees of similarity to the original image. From the visual examples on the left, when the noise is gradually increased to 600, the images’ semantic content compared to the original image changed slowly, and the difference in semantic levels gradually increases. However, when the noise is increased from 600 to 800, the semantic information undergoes a drastic change, leading to a significant difference between the generated image and the original image. The scores given by SeSS are consistent with this trend, demonstrating that SeSS is able to perceive the difference in semantic levels between the images generated by the image generation model and the original image.}
\end{figure*} 
This appendix mainly supplements the performance of SeSS when testing extremely compressed systems, the specific illustration of the test of image generated by Unclip model introducing different noises, and the specific performance of SeSS faced with changes weakly correlated with image semantics. \par
Figure 15 demonstrates the potential of SeSS to evaluate extremely compressed algorithms, as it is able to effectively focus on and assess the degree to which these algorithms preserve key semantic information during extreme compression. This encourages the development of more semantics-based extreme compression algorithms. Figure 16 provides visual examples that supplement the explanation of the reasonable trend of SeSS scores for Unclip-generated images under different noise levels. This proves that SeSS is able to perceive the difference in semantic levels between the images generated by the image generation model and the original image. Figure 17 further demonstrates the strong robustness of SeSS to changes weakly correlated with image semantics and its ability to perceive minor variations. \par
These results further demonstrate the advantages of SeSS in evaluating the outputs of image generation models, and its ability to provide objective evaluation scores in various domains, such as extreme compression. As a result, SeSS is expected to inspire more insightful work in the field of visual semantic communication.
\clearpage

\begin{figure*}[ht]
    \centering
    \includegraphics[scale=0.25]{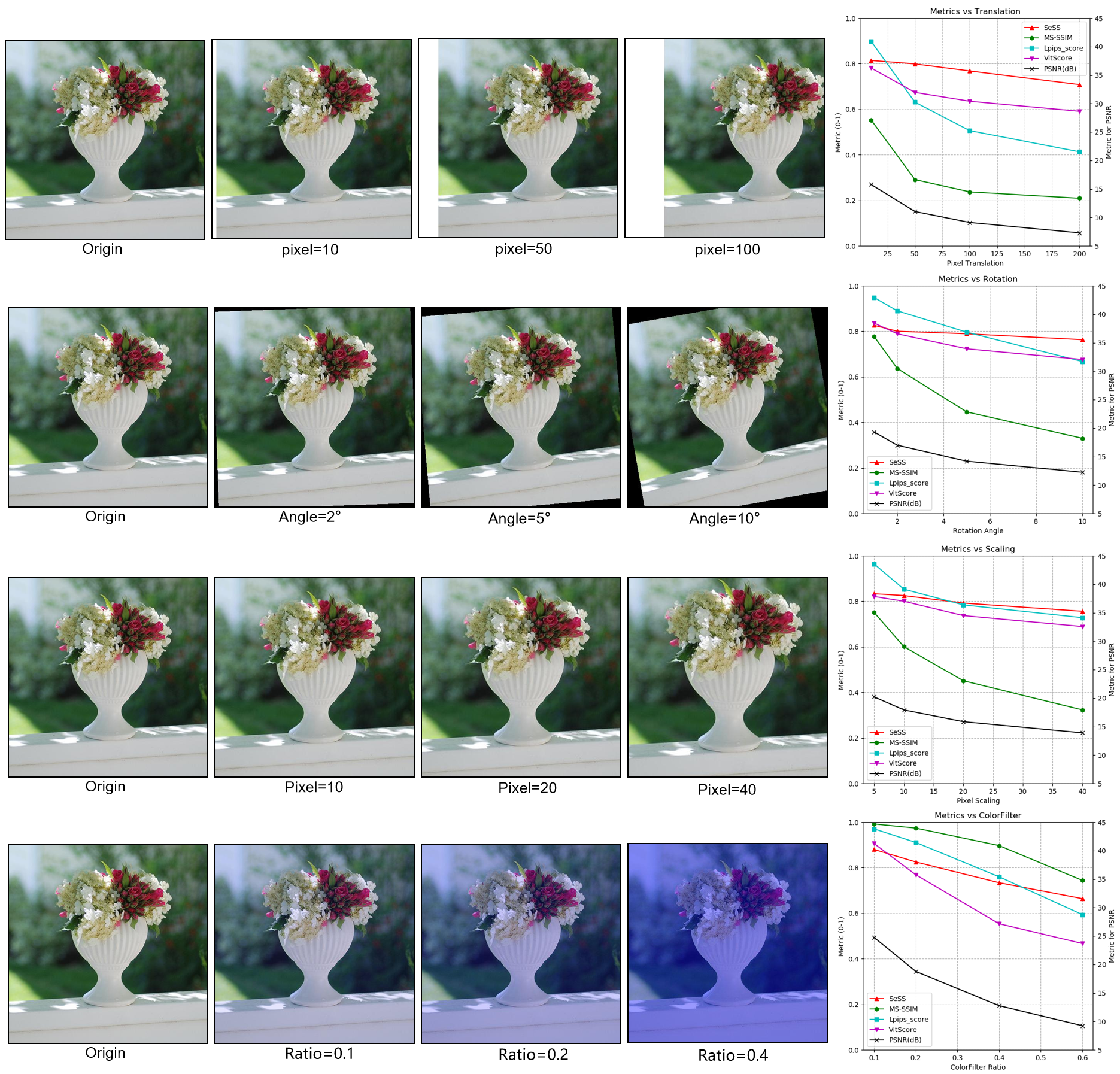}
    \caption{Supplementary explanation for Experiment D. The similarity score of SeSS, MS-SSIM, Lpips\_score, ViTScore, PSNR, on images changed by changes weakly correlated with image semantics. The images are changed with 4 different transformations: (1)Translation: The images are shifted to the left by varying numbers of pixels. (2)Rotation: The images are rotated counter-clockwise by varying degrees. (3) Scaling: The images are cropped from the top, bottom, left, and right by varying numbers of pixels, and then scaled back to the original size.(4) Color Filter: The images are overlaid with a pure blue image at varying blend ratios. SeSS demonstrates extremely strong robustness to these changes, and its metric values decreased smoothly as the degree of transformation increased.}
\end{figure*} 

\section{Acknowledgements}
This work is supported by ....

\bibliographystyle{IEEEtran}
\bibliography{new}

\end{document}